# Multi-Atlas Segmentation with Joint Label Fusion of Osteoporotic Vertebral Compression Fractures on CT

Yinong Wang[1], Jianhua Yao[1], Holger R. Roth[1], Joseph E. Burns[2], and Ronald M. Summers[1]

[1]Imaging Biomarkers and Computer-Aided Diagnosis Laboratory,
Department of Radiology and Imaging Sciences, Clinical Center,
National Institutes of Health, Bethesda, MD 20892
{yinong.wang@nih.gov, jyao@cc.nih.gov, holger.roth@nih.gov,
rsummers@cc.nih.gov}
[2]Department of Radiological Sciences, University of California,
Irvine School of Medicine, Irvine, CA 92697
{jburns@uci.edu}

**Abstract.** The precise and accurate segmentation of the vertebral column is essential in the diagnosis and treatment of various orthopedic, neurological, and oncological traumas and pathologies. Segmentation is especially challenging in the presence of pathology such as vertebral compression fractures. In this paper, we propose a method to produce segmentations for osteoporotic compression fractured vertebrae by applying a multi-atlas joint label fusion technique for clinical CT images. A total of 170 thoracic and lumbar vertebrae were evaluated using atlases from five patients with varying degrees of spinal degeneration. In an osteoporotic cohort of bundled atlases, registration provided an average Dice coefficient and mean absolute surface distance of 92.7±4.5% and 0.32±0.13mm for osteoporotic vertebrae, respectively, and 90.9±3.0% and 0.36±0.11mm for compression fractured vertebrae.

**Keywords.** vertebral compression fracture, vertebra segmentation, multi-atlas label fusion

## 1 Introduction

Vertebral compression fractures (VCFs), which often result from osteoporosis, constitute nearly half the number of clinical osteoporotic fractures. Affecting nearly 1.5 million people in the United States each year, these compression fractures can often result in severe and debilitating pain [1]. Loss in height and the deformity of the vertebral body, with associated changes in the curvature of the spine, can result in a reduction in the spine's effectiveness in weight-bearing, movement, and support. VCFs are typically diagnosed via qualitative visual review, using imaging modalities such as radiography and CT. However, clinical time restrictions prevent detailed processing of data and comprehensive characterization of individual fractures using conventional, manual segmentation methods. Automated methods for identifying and segmenting vertebrae have been applied to the characterization of thoracolumbar spine in patients with non-osteopenia spines [2]. However, it has remained unexplored whether similar applications for osteoporotic patients would be of benefit in the evaluation of cases of trauma and pathology such as vertebral compression fractures.

Accurate segmentation of lumbar and thoracic vertebrae can lead to improved identification and quantitative characterization of VCFs, and have potential to guide fracture treatment and management. Additionally, improved segmentation provides a means for extracting features, such as bone density and anatomic landmarks that allow for a better understanding of the integrity of the spine [3]. VCFs are typically graded manually using Genant scoring, which is based on thresholds for percentage height loss relative to adjacent vertebra [4]. Examples of vertebral compression fractures of different levels of severity are shown in Figure 1.

A number of vertebra segmentation algorithms for computed tomography (CT) images have been proposed. Previous methods that integrated vertebra detection, identification, and segmentation into a single framework were applied primarily to healthy vertebrae [2]. Ibragimov et al. built landmark-based shape representations of vertebrae and aligned models to specific vertebrae in CT images using game theory [5]. An atlas-based method that utilized groupwise segmentation of five vertebrae with majority voting label fusion was proposed in [6] to segment healthy vertebrae in a vertebra segmentation challenge[1] [2]. Fusing information from multiple manually segmented atlases allows for greater acuity to be obtained in the visualization of vertebral compression fractures. In this paper, we present a method to produce segmentations of the osteoporotic spine, especially for vertebrae with compression fractures through a multi-atlas local appearance, similarity weighted joint label fusion technique for clinical CT images [7].

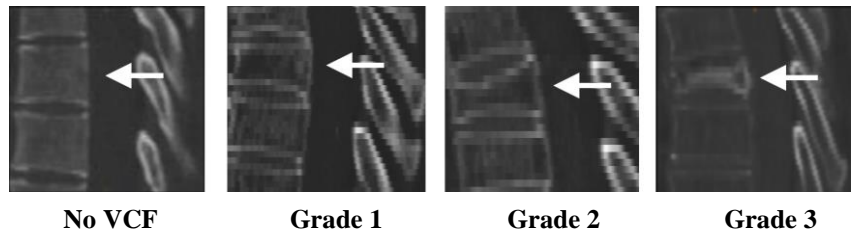

**No VCF**     **Grade 1**     **Grade 2**     **Grade 3**

**Fig. 1.** Sagittal view of vertebrae with varying severities of compression fracture. Arrows point to vertebra of interest. Note increasing osteopenia (decreasing bone density) with higher grades.

## 2 Methods

We propose a multi-atlas joint label fusion framework for spine segmentation on a vertebral basis (Figure 2). Each vertebra was localized by the automatic generation of a bounding box that provided greater local anatomical coherence between images and allowed for more precise image registration. An intensity-based non-rigid B-spline registration technique with affine initialization was used to capture the geometric variability exhibited by vertebrae [8]. Upon completing registration, a set of manually segmented atlases was deformed to the target space using the computed transformations. All transformed atlases were then combined using a joint label fusion technique that utilized local appearance similarity between each registration result to determine the label [7]. Lastly, a morphological label correction step was applied to correct over-segmented results that failed to be properly localized to a single vertebra. This step was used to correct under-segmented results that emerged from the relatively decreased spinal bone density in the osteoporotic test cohort.

---

[1] http://csi-workshop.weebly.com/.

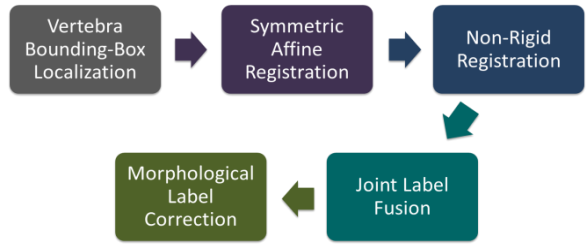

**Fig. 2.** Workflow for our implementation.

### 2.1 Multi-Atlas Registration

The initial location of the vertebra and subsequent bounding box localization was obtained using a fully automatic algorithm reported in [9]. The algorithm is based on adaptive thresholding, watershed, directed graph search, curved reformation, and anatomic vertebra models. The registration was conducted in two stages: symmetric affine transformation for initialization, followed by dense non-rigid registration. Using a floating image $I_2$ from the atlas and a target image $I_1$, a displacement field was computed to estimate an affine transformation $A$ that maximized the intensity similarity between floating and target CT images [8]. The resulting output image $I$ was computed by applying $I_1 \cong I = I_2 \circ A$ where $A$ was then applied to all atlas labels.

After obtaining the initialized affine transformation results, a non-rigid B-spline transformation $T$ was computed that maximized the normalized mutual information ($NMI$) between the floating and target images [8]. $NMI$ between images $I_1$ and $I_2 \circ T$ is computed by using their marginal entropies $H$ and joint entropies $H(I_1, I_2 \circ T)$ as a measure of alignment as

$$NMI = \frac{H(I_1) + H(I_2 \circ T)}{H(I_1, I_2 \circ T)}. \qquad (1)$$

The non-rigid transformation was then optimized on a global and local scale by maximizing the following cost-function

$$C = (1 - \alpha) * NMI - \alpha * P. \qquad (2)$$

The penalty term $P$ uses bending energy to restrict the amount of deformation to achieve physically realistic smooth transformations that penalize only non-rigid registrations [10].

$$P = \frac{1}{N}\sum_{\vec{x}\in\Omega}\left[\left[\left(\frac{\delta^2 T(\vec{x})}{\delta x^2}\right)^2 + \left(\frac{\delta^2 T(\vec{x})}{\delta y^2}\right)^2 + \left(\frac{\delta^2 T(\vec{x})}{\delta z^2}\right)^2\right] + 2*\left[\left(\frac{\delta^2 T(\vec{x})}{\delta xy}\right)^2 + \left(\frac{\delta^2 T(\vec{x})}{\delta yz}\right)^2 + \left(\frac{\delta^2 T(\vec{x})}{\delta xz}\right)^2\right]\right] \quad (3)$$

Weight factor $\alpha$ controls the penalty term; the default value of $\alpha = 0.005$ was used, which has been shown to work well for medical images [8]. Since the registration methods are computationally heavy, a GPU-accelerated solution, NiftyReg[2] was used. Registration produced a transformed set of floating images and segmentation labels which have been warped to the space of the target images.

Two configurations of atlases were explored in our registration experiments: a single vertebra atlas and a three-vertebra atlas, where a vertebra is bundled with the ones directly above and below. T1 and L5 vertebra were bundled with the nearest two thoracic and lumbar vertebrae, respectively. The bundled atlas was a means to preserve the interface between adjacent vertebrae and to prevent conflicting segmentation.

### 2.2 Joint Label Fusion

Since individual registrations to multiple atlases following the affine and non-rigid transformations can yield drastically different results, a joint fusion method (JLF) was applied to the labels in order to enforce region coherence. Since all atlases shared common structures and similar appearances, independent errors in the final segmentation result were minimized following JLF [7]. With a measure of registration success for each transformed atlas, a weighted voting scheme achieved optimal label fusion that minimizes the expectation of combined label differences. Joint label fusion achieves consensus segmentation

$$\bar{S}(x) = \sum_{i=1}^{n} w_i(x)S_i(x) \quad (4)$$

based on individual voting weights $w_i(x)$ and segmentations $S_i(x)$. Weights were determined by

$$w_x = \frac{M_x^{-1} * 1_n}{1_n^t M_x^{-1} 1_n} \quad (5)$$

where $1_n = [1; 1; ...; 1]$ is a vector of size $n$ and $M_x$ is the pairwise dependency matrix that estimates the likelihood of two atlases both producing wrong segmentations on a per-voxel basis for the target image. The end result was a segmentation determined by the probability distribution of propagated labels provided by multi-atlas registrations.

---

[2] http://cmictig.cs.ucl.ac.uk/wiki/index.php/NiftyReg.

### 2.3 Morphological Label Correction

After joint label fusion, post processing measures were taken to correct segmentation error and to refine the boundaries of the binary label result. First, morphological operators and connected components were applied to remove isolated islands and to close holes. Next, collision detection and correction steps were taken to correct segmentations where a voxel was assigned to two vertebrae. A perceptron linear classifier based on the voxel intensity difference and the distance relative to the centers of the vertebrae was applied to determine which vertebra to which the voxel belonged. The weights for intensity difference and distance were set empirically. Finally, a Laplacian level set algorithm was employed to refine the segmentation [11].

## 3 Experiments and Results

### 3.1 Dataset

The dataset used in this experiment was obtained from the University of California, Irvine Medical Center. The dataset contained spine CT data from 10 patients, five from a healthy cohort ranging in age from 20 to 34 years ($27 \pm 5$ yrs), and five from an osteoporotic cohort ranging in age from 59 to 82 years ($73 \pm 9$ yrs) that have been previously identified to have at least one vertebral compression fracture. The studies were performed with spine CT protocol, with in-plane resolution ranging from 0.31 to 0.45mm, and slice thickness ranging from 1mm to 2mm.

A total of 170 thoracic and lumbar vertebrae were evaluated. Among them, 16 vertebrae were previously identified with compression fracture (one with grade 1, ten with grade 2, and five with grade 3). All thoracic and lumbar vertebrae were manually segmented to build atlases for the validation our algorithm. Figure 3 shows reconstructed images from the healthy and osteoporotic data sets and their manual segmentations in a sagittal imaging plane.

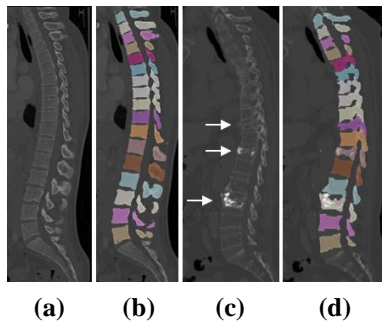

(a)      (b)      (c)      (d)

**Fig. 3.** Sagittal views of spine atlases used. Healthy spine (20 years old) (a) with manual segmentation (b); osteoporotic spine (78 years old) (c) with manual segmentation (d). Arrows point to fractured vertebrae.

### 3.2 Evaluation Methodology

The Dice coefficient (DC) and mean absolute surface distance (ASD) were used for evaluating the accuracy of segmentation. They are defined as

$$DC = \frac{2 * |GT \cap S|}{|GT| + |S|} * 100\% \qquad (6)$$

$$ASD = \frac{1}{|V_S|} \sum_{i=1}^{|V_S|} \|d_i(V_S, V_{GT})\| \qquad (7)$$

where GT and S which refer to the ground truth and the computed segmentations respectively, $V$ refers to the volume, and $d_i$ refers to the distance between the nearest surface voxels of each label [12]. The vertebrae were partitioned into four substructures using the algorithm in [3]: vertebral body (VB), left transverse process (LTP), right transverse process (RTP) and spinous process (SP). Performance was evaluated on both the whole vertebra and the vertebral body substructure.

Performance was evaluated for either using healthy or osteoporotic cases as atlases. We compared three experimental setups: (1) use of healthy vertebrae as atlases to segment the osteoporotic vertebrae with compression fractures (H2D) (2) use of the osteoporotic vertebrae as atlases for segmentation of osteoporotic vertebrae (D2D) and (3) use of a three-vertebra bundled atlas using osteoporotic data (BD2D). A leave-one-vertebra-out scheme was adopted in the second and third experiments.

### 3.3 Results

Experiments were conducted to assess the segmentation of healthy and osteoporotic vertebrae and vertebrae with compression fracture. Figure 4 shows the visual segmentation result at each step of the process. Individual registrations using GPU-based NiftyReg were completed with a runtime of 5 minutes each, with an additional 5 minutes for the label fusion [8]. Table 1 lists the characteristics (volume and density) of osteoporotic vertebrae at different grades of compression fracture and the segmentation performance in different experimental setups.

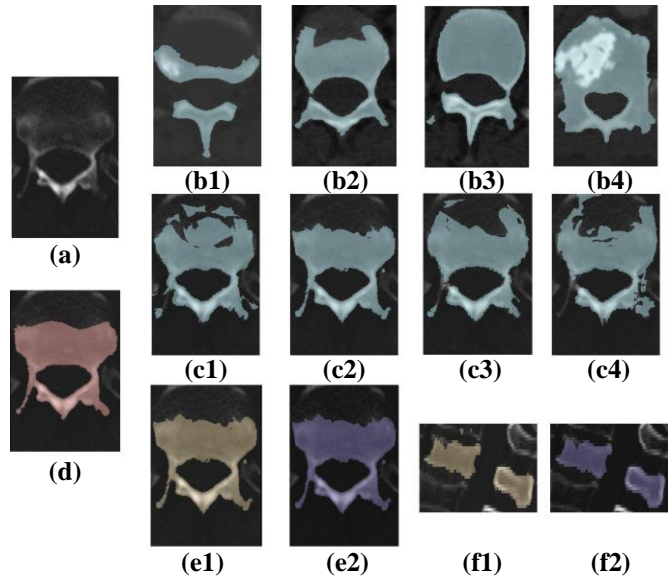

**Fig. 4.** Segmentation results at each step of the process. Original image (a), atlas labels (b1-b4), registration results (c1-c4), ground truth segmentation (d), and joint label fusion (e1, f1) and corrected result (e2, f2) in axial and sagittal views.

BD2D setup performed better than both H2D and D2D especially for fractured vertebrae as shown in Table 1, suggesting that the bundled osteoporotic atlases provided the best registration results. However, the performance on grade 3 compression fractures was not on the same level as those of grades 1 and 2. A paired t-test showed statistically significant improvement in the Dice coefficient and mean absolute surface distance when using BD2D instead of H2D atlas in all metrics of segmentation performance ($p < 0.001$).

Table 2 shows the segmentation performance on both the whole vertebra and the vertebral body using the three groups of atlases before and after morphological label correction. While the Dice coefficient was higher for the whole vertebra for all atlas registration groups, the mean absolute surface distance performed better on just the vertebral body substructure. Depending on the atlas and the segmentation performance metric used, the morphological label correction improved results over the label fusion alone. However, using BD2D, the improvements seen in the Dice coefficient and mean absolute surface distance in both the whole vertebra and vertebra body substructure were statistically significant ($p < 0.05$).

**Table 1.** Comparison of upper thoracic (UT), lower thoracic (LT), lumbar (L), non-fracture (NF), grades 1-3 (G1, G2, G3) compression fracture, VCF, and all vertebrae. Mean and SD (in parenthesis) of segmentation volume (cm$^3$), density (HU), DC (%) and ASD (mm) of refined (-r) H2D, D2D, and BD2D setups are reported. Best performing methods are bolded.

|     | No. Vert. | Vol. | Den. | DC H2D-r | DC D2D-r | DC BD2D-r | ASD H2D-r | ASD D2D-r | ASD BD2D-r |
|-----|-----------|------|------|----------|----------|-----------|-----------|-----------|------------|
| **UT** | 30 | 21 (3.7) | 259 (48) | 90.6 (4.7) | 91.3 (4.9) | 92.5 (1.5) | 0.33 (0.16) | 0.32 (0.15) | 0.29 (0.06) |
| **LT** | 30 | 33 (6.7) | 274 (96) | 92.4 (3.6) | 91.9 (3.4) | 93.0 (2.8) | 0.34 (0.10) | 0.37 (0.13) | 0.34 (0.10) |
| **L** | 25 | 56 (8.1) | 267 (86) | 91.5 (10.5) | 93.5 (4.5) | 92.5 (7.7) | 0.42 (0.40) | 0.37 (0.35) | 0.35 (0.21) |
| **NF** | 69 | 36 (16) | 254 (52) | 92.0 (7.1) | 92.6 (4.4) | 93.0 (4.7) | 0.35 (0.27) | 0.34 (0.24) | 0.31 (0.14) |
| **G1** | 1 | 22 | 216 | 88.3 | 92.4 | 92.2 | 0.31 | 0.34 | 0.39 |
| **G2** | 10 | 31 (11) | 259 (41) | 91.0 (2.7) | 92.3 (1.3) | 92.6 (1.1) | 0.32 (0.07) | 0.30 (0.05) | 0.29 (0.04) |
| **G3** | 5 | 39 (16) | 465 (160) | 86.2 (3.8) | 86.1 (3.0) | 87.6 (3.6) | 0.55 (0.09) | 0.53 (0.08) | 0.50 (0.08) |
| **VCF** | 16 | 33 (12) | 321 (134) | 89.3 (3.7) | 90.4 (3.5) | **91.0 (3.2)** | 0.39 (0.13) | 0.37 (0.12) | **0.36 (0.11)** |
| **All** | 85 | 35 (15) | 267 (78) | 91.5 (6.6) | 92.1 (4.3) | **92.7 (4.5)** | 0.36 (0.25) | 0.35 (0.23) | **0.32 (0.13)** |

**Table 2.** Mean and SD of DC (%) and ASD (mm) for whole vertebra and vertebral body (VB) using JLF of H2D, D2D, BD2D and refined results. Two statistical tests were conducted: comparison of (1) each atlas type with its refined (-r) segmentations and (2) H2D-r and D2D-r with BD2D-r. Varying levels of significance in the first test are marked by asterisks, where * indicates $p \leq 0.05$, ** $p \leq 0.01$, and *** $p \leq 0.001$. Significance of $p \leq 0.05$ in the second test is indicated by §.

|        | DC | DC_VB | ASD | ASD_VB |
|--------|----|----|----|----|
| **H2D** | 91.2 (6.8) | 88.0 (21.1) | 0.35 (0.17) | 0.33 (0.14) |
| **H2D-r** | 91.5$^*$ (6.6) | 88.7$^{**}$ (18.3) | 0.36 (0.25) | 0.32 (0.11) |
| **D2D** | 92.0 (4.4) | 89.2 (17.7) | 0.33 (0.16) | 0.31 (0.14) |
| **D2D-r** | 92.1 (4.3) | 89.4$^{***}$ (17.7) | 0.35 (0.23) | 0.32$^{***}$ (0.15) |
| **BD2D** | 92.5 (4.4) | 89.5 (17.9) | 0.31 (0.10) | 0.30 (0.09) |
| **BD2D-r** | 92.7$^{**\S}$ (4.5) | 89.7$^{**\S}$ (18.0) | 0.32$^{*\S}$ (0.13) | 0.30$^{***\S}$ (0.10) |

## 4 Discussion

The multi-atlas joint label fusion segmentation method presented in this paper provides a robust framework for the segmentation and analysis of osteoporotic spine and vertebral compression fractures. Compared to prior literature which used multi-atlas joint label fusion techniques for healthy spine [2], our method produced similar or better performance, as measured by Dice coefficient, for both non-fractured osteopo-rotic vertebrae (93.0%) and compression fractured vertebrae (91.0%). A similar pattern was observed in the mean absolute surface distance where we obtained a mean ASD of 0.31mm and 0.36mm for non-fractured and fractured vertebrae, respectively. The same five osteoporotic cases were re-evaluated by the techniques used in [13]. The two best performing methods, one of which was not fully-automated and used statistical shape model deformation and B-spline relaxation [14], and one similar to ours that used multi-atlas registrations on bundles of five vertebrae with basic majority voting label fusion [6], produced similar overall DC of 89.8% and 89.7% and ASD of 0.64mm and 0.86mm respectively. Our technique, which utilized multi-atlas registrations on bundles of three vertebrae with joint label fusion based on local similarity appearance with a weighting scheme and additional morphological label corrections, performed at a DC of 92.7% and ASD of 0.32mm.

Certain metrics in our algorithm can be selected and further optimized to improve performance. In image registration, the local normalized correlation coefficient (LNCC) could be used instead of normalized mutual information (NMI) within NiftyReg [8], assuming that a linear relationship exists between the intensities of the voxels of the floating and target images. The weight factor $\alpha$ can be varied to determine the best value to use on CT images, as the default value was optimized for inter-modality registrations of medical images. In addition, the cases chosen for these experiments may not have fully captured the variability of morphology in the osteoporotic population. More cases and optimized atlas selection could produce a segmentation algorithm than can be applied to a more robust population of osteoporotic individuals.

Based on results in Tables 1 and 2, segmentation performance on grade 3 compression fractures was not as high as those on grades 1 and 2. One potential reason for this is the markedly elevated intensities observed in the grade 3 VCFs as a result of injected surgical cement in three of the five fractured vertebrae. Additional fractured vertebrae are necessary to increase and test the robustness of the algorithm, as there were only 16 VCFs total, of which only one grade 1 in the test cohort.

The average Dice coefficient was higher when evaluating the whole vertebrae than on a single vertebral body. However, the opposite was true when examining the mean absolute surface distance. This is likely the result of evaluating smaller and less variable regions at the surface of the vertebral body rather than more variable regions across the entire vertebra.

## 5 Conclusion

In this paper, we present a robust method for segmenting osteoporotic compression fractures. Despite significantly greater variability in the morphology of compression fractured spines, we have shown that optimization of existing techniques for non-osteopenia spines can yield similar or better performance for segmentation of osteoporotic spines. With these improved segmentations, morphological features of osteoporotic compression fractured vertebrae can be better characterized for clinical appli-cations. The extraction of features through an improved segmentation result may also provide means for better treatment and management of spinal disease.

## Acknowledgments

This research was supported in part by the Intramural Research Program of National Institutes of Health Clinical Center.

## References

1. B. L. Riggs and L. J. Melton, III, "The worldwide problem of osteoporosis: insights afforded by epidemiology," *Bone,* vol. 17, pp. 505S-511S, Nov 1995.
2. J. Yao and S. Li, "Report of Vertebra Segmentation Challenge in 2014 MICCAI Workshop on Computational Spine Imaging," in *Recent Advances in Computational Methods and Clinical Applications for Spine Imaging: Lecture Notes in Computational Vision and Biomechanics*. vol. 20, J. Yao, B. Glocker, T. Klinder, and S. Li, Eds., ed: Springer International Publishing, 2015, pp. 247-259.
3. J. Yao, J. E. Burns, S. Getty, J. Stieger, and R. M. Summers, "Automated extraction of anatomic landmarks on vertebrae based on anatomic knowledge and geometrical constraints," in *2014 IEEE 11th International Symposium on Biomedical Imaging (ISBI)*, 2014, pp. 397-400.
4. H. K. Genant, C. Y. Wu, C. Van Kuijk, and M. C. Nevitt, "Vertebral fracture assessment using a semiquantitative technique," *Journal of Bone and Mineral Research,* vol. 8, pp. 1137-1148, 1993.
5. B. Ibragimov, B. Likar, F. Pernus, and T. Vrtovec, "Shape representation for efficient landmark-based segmentation in 3-d," *IEEE Transactions on Medical Imaging,* vol. 33, pp. 861-74, Apr 2014.
6. D. Forsberg, "Atlas-based segmentation of the thoracic and lumbar vertebrae," in *Recent Advances in Computational Methods and Clinical Applications for Spine Imaging*, ed: Springer, 2015, pp. 215-220.
7. H. Wang, J. W. Suh, S. R. Das, J. B. Pluta, C. Craige, and P. Yushkevich, "Multi-atlas segmentation with joint label fusion," *IEEE Transactions on Pattern Analysis and Machine Intelligence,* vol. 35, pp. 611-623, 2013.
8. M. Modat, G. R. Ridgway, Z. A. Taylor, M. Lehmann, J. Barnes, D. J. Hawkes, *et al.*, "Fast free-form deformation using graphics processing units," *Computer Methods and Programs in Biomedicine,* vol. 98, pp. 278-284, 2010.
9. J. Yao, S. D. O'Connor, and R. M. Summers, "Automated spinal column extraction and partitioning," in *3rd IEEE International Symposium on Biomedical Imaging: Nano to Macro, 2006.*, 2006, pp. 390-393.
10. D. Rueckert, L. I. Sonoda, C. Hayes, D. L. G. Hill, M. O. Leach, and D. J. Hawkes, "Nonrigid Registration Using Free-Form Deformations: Application to Breast MR Images," *IEEE Transactions on Medical Imaging,* vol. 18, pp. 712-721, 1999.

11. H. Hui and J. Jionghui, "Laplacian Operator Based Level Set Segmentation Algorithm for Medical Images," in *CISP '09. 2nd International Congress on Image and Signal Processing, 2009.*, 2009, pp. 1-5.
12. L. R. Dice, "Measures of the amount of ecologic association between species," *Ecology,* vol. 26, pp. 297-302, 1945.
13. J. Yao, J. E. Burns, D. Forsberg, A. Seitel, A. Rasoulian, P. Abolmaesumi*, et al.*, "Comparison and evaluation of vertebra segmentation methods for CT images: the CSI 2014 challenge," *Computerized Medical Imaging and Graphics,* 2015.
14. I. Castro-Mateos, J. M. Pozo, A. Lazary, and A. Frangi, "3D Vertebra Segmentation by Feature Selection Active Shape Model," in *Recent Advances in Computational Methods and Clinical Applications for Spine Imaging*. vol. 20, J. Yao, B. Glocker, T. Klinder, and S. Li, Eds., ed: Springer International Publishing, 2015, pp. 241-245.